\documentclass[10pt,journal,compsoc]{IEEEtran}
\ifCLASSOPTIONcompsoc
  \usepackage[nocompress]{cite}
\else
  \usepackage{cite}
\fi

 \usepackage[dvips]{graphicx}

\usepackage{amscd,amssymb,mathrsfs,amsmath,eqparbox,array}
\newtheorem{thm}{Theorem}[section]

\newtheorem{lemma}[thm]{Lemma}

\newtheorem{defin}[thm]{Definition}

\numberwithin{equation}{section} 

\def\R {{\mathbb{R}}}

\def\N {{\mathbb{N}}}

\begin{document}
%
\title{Function space analysis of deep learning representation layers}
%
%
%
%

\author{Oren~Elisha
        and~Shai~Dekel
\IEEEcompsocitemizethanks{\IEEEcompsocthanksitem O. Elisha is with GE Research
and the School of Mathematics, Tel-Aviv University, Tel-Aviv \protect\\

\IEEEcompsocthanksitem S. Dekel is with WIX AI and the School of Mathematics, Tel-Aviv University, Tel-Aviv}}

\IEEEtitleabstractindextext{%
\begin{abstract}
In this paper we propose a function space approach to Representation Learning \cite{Ben} and the analysis of the representation layers in deep learning architectures. We show how to compute a `weak-type'  Besov smoothness index that quantifies the geometry of the clustering in the feature space. This approach was already applied successfully to improve the performance of machine learning algorithms such as the Random Forest \cite{ED} and tree-based Gradient Boosting \cite{DEM}. Our experiments demonstrate that in well-known and well-performing trained  networks, the Besov smoothness of the training set, measured in the corresponding hidden layer feature map representation, increases from layer to layer. We also contribute to the understanding of generalization \cite{Zhang} by showing how the Besov smoothness of the representations, decreases as we add more mis-labeling to the training data. We hope this approach will contribute to the de-mystification of some aspects of deep learning.  
\end{abstract}

}

\maketitle

\IEEEdisplaynontitleabstractindextext

%
\IEEEpeerreviewmaketitle

\IEEEraisesectionheading{\section{Introduction}\label{sec:introduction}}

%
%
%
%
\IEEEPARstart{A}{n} excellent starting point for this paper is the survey on Representation Learning \cite{Ben}. One of the main issues raised in this survey is that simple smoothness assumptions on the data do not hold. That is, there exists a curse of dimensionality and `close' feature 
representations do not map to `similar' values. The authors write: ``We advocate learning algorithms that are flexible and non-parametric but do not rely exclusively on the smoothness assumption". 

In this work we do in fact advocate smoothness analysis of representation 
layers, yet in line with \cite{Ben}, our notion of smoothness is indeed flexible, adaptive and non-parametric. 
We rely on geometric multivariate function space theory and use a machinery of Besov `weak-type' smoothness which is robust enough to support quantifying smoothness of high dimensional discontinuous functions.  

Although machine learning is mostly associated with the field of statistics we argue that 
the popular machine learning algorithms such as Support Vector Machines, Tree-Based Gradient Boosting and
Random Forest (see e.g \cite{HTF}), are in fact closely related to the field of multivariate adaptive approximation theory. 
In essence, these algorithms work best if there exists geometric structure of clusters in the feature space.
If such geometry exists, these algorithms will capture it, by segmenting out the different clusters.
We claim that in the absence of such  geometry, these machine learning algorithms will fail. 

However, this is exactly where Deep Learning (DL) comes into play. In the absence of a geometrical structure in the given
initial representation space, the goal of the DL layers is to create a series of transformations from one representation space
to the next, where structure of the geometry of the clusters improves sequentially. We quote \cite{Ch}: ``The whole process of applying this complex 
geometric transformation to the input data can be visualized in 3D by imagining a person trying to uncrumple a paper ball: the crumpled 
paper ball is the manifold of the input data that the model starts with. Each movement operated by the person on the 
paper ball is similar to a simple geometric transformation operated by one layer. The full uncrumpling gesture sequence 
is the complex transformation of the entire model. Deep learning models are mathematical machines for uncrumpling complicated manifolds 
of high-dimensional data''. 

Let us provide an instructive example:
Assume we are presented with a set of gray-scale images of dimension $\sqrt{n_0} \times \sqrt{n_0}$ with $L$ class labels. Assume further that 
a DL network has been successfully trained to classify these images with relatively high precision. This allows
us to extract the representation of each image in each of the hidden layers. To create a representation at layer $0$, we concatenate the $\sqrt{n_0}$
rows of pixel values of each image, to create a vector of dimension $n_0$. We also normalize the pixel
values to the range $[0,1]$. Since we advocate a function-theoretical approach,
we transform the class labels into vector-values in the space $\R^{L-1}$ by assigning 
each label to a vertex of a standard simplex (see Section \ref{sec:overview} below). Thus, the images are considered as samples of a function
$f_0:[0,1]^{n_0} \rightarrow \R^{L-1}$. In the general case, there is no hope that there exists geometric 
clustering of the classes in this initial feature space and that $f_0$ has sufficient `weak-type' smoothness (as is verified by our experiments below). 
Thus, a transform into a different
feature space is needed. We thus associate with each $k$-th layer of a DL network, a function 
$f_k:[0,1]^{n_k} \rightarrow \R^{L-1}$ where the samples are vectors created by normalizing and concatenating the feature maps
computed from each of the images. 
Interestingly enough, although the series of functions $f_k$ are embedded in different dimensions $n_k$,
through the simple normalizing of the features, our method is able to assign smoothness indices to each layer that are comparable. We claim that for well
performing networks, the representations in general `improve' from layer to layer and that our method captures this phenomena and shows the increase of smooothness. 

Related work is \cite{PRE} where an architecture of Convolutional Sparse Coding was analyzed. The connection to this work is the emphasis on `sparsity' analysis of hidden layers. However, there are significant differences since we advocate a function space theoretical analysis of any neural network architecture in current use. Also, there is the recent work \cite{ST}, where the authors take an `information-theoretical' approach
to the analysis of the stochastic gradient descent optimization of DL networks and the representation layers. One can safely say that all
of the approaches, including the one presented here, need to be further evaluated on larger datasets and deeper architectures. 

The paper is organized as follows: In Section \ref{sec:overview} we review our smoothness analysis machinery which 
is the Wavelet Decomposition of Random Forest (RF)  \cite{ED}. In Section \ref{sec:mylabel2} we 
present the required geometric function space theoretical background. Since we are comparing different representations over different spaces of different dimensions, 
we add to the theory presented in \cite{ED} relevant `dimension-free' results. In Section \ref{sec:smoothAnalysis} we show how to apply the theory in practice. 
Specifically, how the wavelet decomposition of a RF can be used to 
numerically compute a Besov `weak-type' smoothness index of a given function in any representation space (e.g hidden layer).
Section \ref{sec:applications} provides experimental results that 
demonstrate how our theory is able to explain empirical findings in various scenarios.
Finally, Section \ref{sec:conc} presents our conclusions as well as future work. 

\section{Wavelet decomposition of Random Forests}\label{sec:overview}

To measure smoothness of a dataset at the various DL representation layers, we apply the construction of wavelet 
decompositions of Random Forests \cite{ED}. Wavelets \cite{Da}, 
\cite{Ma} and geometric wavelets \cite{DL}, \cite{AAD}, are a 
powerful yet simple tool for constructing sparse representations of `complex' 
functions. The Random Forest (RF) \cite{BS}, \cite{CSK}, \cite{HTF} introduced by Breiman \cite{Br1}, \cite{Br2}, is a 
very effective machine learning method that can be considered as a way to 
overcome the `greedy' nature and high variance of a single decision tree. 
When combined, the wavelet decomposition of the RF unravels the sparsity of 
the underlying function and establishes an order of the RF nodes from 
`important' components to `negligible' noise. Therefore, the method provides 
a better understanding of any constructed RF. Furthermore, the method is a basis for a robust
feature importance algorithm. We note that one can apply a similar approach to improve the performance of tree-based Gradient Boosting algorithms \cite{DEM}.

We begin with an overview of single trees. In statistics and machine 
learning \cite{BFSO}, \cite{Alp}, \cite{BS}, \cite{DMF}, \cite{HTF} the 
construction is called a Decision Tree or the Classification and Regression 
Tree (CART). We are given a real-valued 
function $f\in L_{2} \left( {\Omega_{0} } \right)$ or a discrete dataset 
$\left\{ {x_{i} \in \Omega_{0} ,f\left( {x_{i} } \right)} \right\}_{i\in I} 
$, in some convex bounded domain $\Omega_{0} \subset \R^{n}$. The goal 
is to find an efficient representation of the underlying function, overcoming the 
complexity, geometry and possibly non-smooth nature of the function values. 
To this end, we subdivide the initial domain $\Omega_{0} $ into two 
subdomains, e.g. by intersecting it with a hyper-plane. The subdivision is 
performed to minimize a given cost function. This subdivision process then 
continues recursively on the subdomains until some stopping criterion is met, 
which in turn, determines the leaves of the tree. We now describe one 
instance of the cost function which is related to minimizing variance. 
At each stage of the subdivision process, at 
a certain node of the tree, the algorithm finds, for the convex domain 
$\Omega \subset {\R}^{n}$ associated with the node:

\noindent (i) A partition by an hyper-plane into two convex subdomains ${\Omega }',{\Omega }''$, $\Omega'\cup {\Omega }''=\Omega$, 

\noindent (ii) Two multivariate polynomials $Q_{{\Omega }'} ,Q_{{\Omega }''} \in \Pi_{r-1} \left( {\R^{n}} \right)$, 
of fixed (typically low) total degree r $-$ 1.

The partition and the polynomials are chosen to minimize the following quantity
\begin{equation}
\label{eq1}
\left\| {f-Q_{\Omega' } } \right\|_{L_{p} \left( {{\Omega }'} 
\right)}^{p} +\left\| {f-Q_{\Omega'' } } \right\|_{L_{p} \left( {{\Omega 
}''} \right)}^{p}.
\end{equation}
Here, for $1 \le p < \infty$, we used the definition 
\[
\left\| g \right\|_{L_{p} \left( {\tilde{{\Omega }}} \right)} :=\left( 
{\int_{\tilde{{\Omega }}} {\left| {g\left( x \right)} \right|^{p}dx} } 
\right)^{1/p}.
\]
If the dataset is discrete, consisting of feature vectors $x_{i} \in \R^{n},i\in I$ , 
with response values $f\left( {x_{i} } \right)$, 
then a discrete functional is minimized over all partitions $\Omega'\cup {\Omega }''=\Omega$ 
\begin{equation}
\label{dis-eq1}
\sum\limits_{x_{i} \in {\Omega }'} {\left| {f\left( {x_{i} } 
\right)-Q_{{\Omega }'}(x_i) } \right|^{p}} +\sum\limits_{x_{i} \in {\Omega }''} 
{\left| {f\left( {x_{i} } \right)-Q_{{\Omega }''}(x_i) } \right|^{p}}.
\end{equation}

Observe that for any given subdividing hyperplane, the approximating 
polynomials in (\ref{dis-eq1}) can be uniquely determined for $p=2$, by least square 
minimization. For the order $r=1$, the approximating polynomials are nothing 
but the mean of the function values over each of the subdomains
\begin{eqnarray*} \label{C_omega}
Q_{{\Omega }'} \left( x \right)=C_{{\Omega }'} =\frac{1}{\# \left\{ {x_{i} 
\in {\Omega }'} \right\}}\sum\limits_{x_{i} \in {\Omega }'} {f\left( {x_{i} 
} \right)} , \\
Q_{{\Omega }''} \left( x \right)=C_{{\Omega }''} =\frac{1}{\# \left\{ {x_{i} 
\in {\Omega }''} \right\}}\sum\limits_{x_{i} \in {\Omega }''} {f\left( 
{x_{i} } \right)} .
\end{eqnarray*}
In many applications of decision trees, the high-dimensionality of the data 
does not allow to search through all possible subdivisions. As in our 
experimental results, one may restrict the subdivisions to the class of 
hyperplanes aligned with the main axes. In contrast, there are cases where 
one would like to consider more advanced form of subdivisions, where they 
take certain hyper-surface form or even non-linear forms through kernel Support Vector
Machines. Our paradigm of 
wavelet decompositions can support in principle all of these forms.

Random Forest (RF) is a popular machine learning tool that collects decision 
trees into an ensemble model \cite{Br1},\cite{BS}. The trees are 
constructed independently in a diverse fashion and prediction is done by a 
voting mechanism among all trees. A key element \cite{Br1}, is that large diversity between the trees 
reduces the ensemble's variance. There are many RFs variations that differ 
in the way randomness is injected into the model, e.g bagging, random 
feature subset selection and the partition criterion \cite{CSK}, 
\cite{HTF}. Our wavelet decomposition paradigm is 
applicable to most of the RF versions known from the literature.

Bagging \cite{Br2} is a method that produces partial 
replicates of the training data for each tree. A typical approach is to 
randomly select for each tree a certain percentage of the training set (e.g. 
80{\%}) or to randomly select samples with repetitions 
\cite{HTF}. 

Additional methods to inject randomness can be achieved at the node 
partitioning level. For each node, we may restrict the partition criteria to 
a small random subset of the parameter values (hyper-parameter). A typical 
selection is to search for a partition from a random subset of $\sqrt n$~  
features \cite{Br1}. This technique is also useful for 
reducing the amount of computations when searching the appropriate partition 
for each node. Bagging and random feature selections are not mutually 
exclusive and could be used together. 

For $j=1,...,J$, one creates a decision tree ${\cal T}_{j} $, 
based on a subset of the data, $X^{j}$. One then provides a weight (score) 
$w_{j} $ to the tree ${\cal T}_{j} $, based on the estimated performance of 
the tree, where $\sum_{j=1}^J{w_j}=1$. In the supervised learning, one typically uses the remaining data 
points $x_{i} \notin X^{j}$ to evaluate the performance of ${\cal T}_{j} $. 
For simplicity, we will mostly consider in this paper the choice of uniform 
weights $w_{j} =1/J$. For any point $x\in \Omega_{0} $, the approximation associated with 
the $j^{th}$ tree, denoted by $\tilde{{f}}_{j} \left( {x} \right)$, is 
computed by finding the leaf $\Omega \in {\cal T}_{j} $ in which $x$ is 
contained and then evaluating $\tilde{{f}}_{j} \left( {x_{i} } 
\right):=Q_{\Omega } \left( {x} \right)$, where $Q_{\Omega } $ is the 
corresponding polynomial associated with the decision node $\Omega $. 
One then assigns an approximate value to any point $x\in \Omega_{0} $ by
\[
\tilde{{f}}\left( x \right)=\sum\limits_{j=1}^J {w_{j} \tilde{{f}}_{j} 
\left( x \right)}.
\]

Typically, in classification problems, the response variable does have a 
numeric value, but is labeled by one of L classes. In this scenario, 
each input training point $x_{i} \in \Omega_{0}$ is assigned with a class 
$Cl\left( {x_{i} } \right)$. To convert the problem to the `functional' 
setting described above one assigns to each class the value of a node on 
the regular simplex consisting of $L$ vertices in ${\R}^{L-1}$ (all 
with equal pairwise distances). Thus, we may assume that the input data is 
in the form
\[
\left\{ {x_{i} ,Cl\left( {x_{i} } \right)} \right\}_{i\in I} \in \left( {\R^{n},{\R}^{L-1}} \right).
\]
In this case, if we choose approximation using constants ($r=1)$, then the 
calculated mean over any subdomain $\Omega $ is in fact a point 
$\vec{{E}}_{\Omega } \in {\R}^{L-1}$, inside the simplex. Obviously, any 
value inside the multidimensional simplex, can be mapped back to a class, 
along with an estimated confidence level, by calculating the closest vertex 
of the simplex to it. 

Following the classic 
paradigm of nonlinear approximation using wavelets 
\cite{Da},\cite{De},\cite{Ma} and the geometric function space theory presented in \cite{DL}, \cite{KP}, 
we introduced in \cite{ED} a construction of a wavelet 
decomposition of a forest. Let ${\Omega }'$ be a child of $\Omega $ in a 
tree ${\cal T}$, i.e. ${\Omega }'\subset \Omega $ and ${\Omega }'$ was 
created by a partition of $\Omega $. 
Denote by ${\rm {\bf 1}}_{{\Omega }'} $, the indicator function over the 
child domain ${\Omega }'$, i.e. ${\rm {\bf 1}}_{{\Omega }'} \left( x 
\right)=1$, if $x\in {\Omega }'$ and ${\rm {\bf 1}}_{{\Omega }'} \left( x 
\right)=0$, if $x\notin {\Omega }'$. We use the polynomial approximations 
$Q_{{\Omega }'} ,Q_{\Omega } \in \Pi_{r-1} \left( {{\R}^{n}} \right)$, 
computed by the local minimization (\ref{eq1}) and define
\begin{equation}
\label{eq3}
\psi_{{\Omega }'}(x) :=\psi_{{\Omega }'} \left( f \right)(x):={\rm {\bf 
1}}_{{\Omega }'}(x) \left( {Q_{{\Omega }'}(x) -Q_{\Omega } }(x) \right),
\end{equation}
as the\textbf{ geometric wavelet} associated with the subdomain ${\Omega }'$ 
and the function $f$, or the given discrete dataset $\left\{ {x_{i} ,f\left( 
{x_{i} } \right)} \right\}_{i\in I} $. Each wavelet $\psi_{{\Omega }'} $, 
is a `local difference' component that belongs to the detail space between 
two levels in the tree, a `low resolution' level associated with $\Omega $ 
and a `high resolution' level associated with ${\Omega }'$. Also, the 
wavelets (\ref{eq3}) have the `zero moments' property, i.e., if the response variable is 
sampled from a polynomial of degree $r-1$ over $\Omega $, then our local 
scheme will compute $Q_{{\Omega }'} \left( x \right)=Q_{\Omega } \left( x 
\right)=f\left( x \right)$, $\forall x\in \Omega $, and therefore $\psi 
_{{\Omega }'} =0$. 

Under certain mild conditions on the tree ${\cal T}$ and the function $f$, 
we have by the nature of the wavelets, the `telescopic' sum of differences

\begin{equation}
\label{eq3a}
f=\sum\limits_{\Omega \in {\cal T}} {\psi_{\Omega } } , \qquad \psi 
_{\Omega_{0} } :=Q_{\Omega_{0} }.
\end{equation}
For example, (\ref{eq3a}) holds in $L_{p} $-sense, $1\le p<\infty $, if $f\in L_{p} 
\left( {\Omega_{0} } \right)$ and for any $x\in \Omega_{0} $ and series of 
domains $\Omega_{l} \in {\cal T}$, each on a level $l$, with $x\in \Omega 
_{l} $ , we have that $\lim\limits_{l\to \infty } \mbox{diam}\left( 
{\Omega_{l} } \right)=0$.

The norm of a wavelet is computed by
\[
\left\| {\psi_{{\Omega }'} } \right\|_{p}^{p} =\int_{{\Omega }'} {\left( 
{Q_{{\Omega }'} \left( x \right)-Q_{\Omega } \left( x \right)} 
\right)^{p}dx}.
\]
For the case $r=1$, where $Q_{{\Omega }} \left( x \right)=C_{\Omega}$ and $Q_{{\Omega }'} \left( x \right)=C_{\Omega'}$ this simplifies to 
\begin{equation}
\label{eq4}
\left\| {\psi_{{\Omega }'} } \right\|_{p}^{p} =\left| {{C}_{{\Omega }'} -C_{\Omega 
} } \right|^{p} \left| \Omega' \right|,
\end{equation}
where  $\left| \Omega' \right|$ denotes the volume of $\Omega'$. 
Observe that for $r=1$, the subdivision process for partitioning a node by minimizing 
(\ref{eq1}) is equivalent to maximizing the sum of squared norms of the wavelets that 
are formed in that partition (see \cite{ED}).

Recall that our approach is to convert classification problems into a `functional' setting
by assigning the $L$ class labels to vertices of a simplex in $\R^{L-1}$. In such cases of multi-valued 
functions, choosing $r=1$, the wavelet $\psi_{{\Omega }'} :{\R}^{n}\to \R^{L-1}$ is 
\[
\psi_{{\Omega }'}(x) ={\rm {\bf 1}}_{{\Omega }'}(x) \left( {\vec{{E}}_{{\Omega 
}'} -\vec{{E}}_{\Omega } } \right),
\]
and its norm is given by 
\begin{equation}
\label{eq5}
\left\| {\psi_{{\Omega }'} } \right\|_{p}^{p} =\left\| {\vec{{E}}_{{\Omega }'} -\vec{{E}}_{\Omega 
} } \right\|_{l_{2} }^{p} \left| \Omega' \right|,
\end{equation}
where for $\vec{{v}}\in {\R}^{L-1}$,$\left\| {\vec{{v}}} \right\|_{l_{2} 
} :=\sqrt {\sum\nolimits_{i=1}^{L-1} {v_{i}^{2}} } $. 

\begin{figure}
\centering
\includegraphics[width=3in]{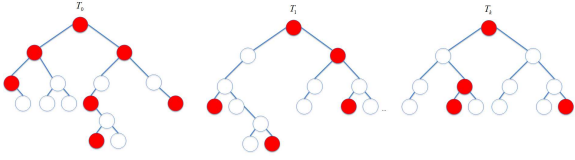}
\caption{Selection of an M-term approximation from the entire forest. }
\label{fig2}
\end{figure}

Using any given weights assigned to the trees, we obtain a wavelet representation of the entire RF
\begin{equation}
\label{eq6}
\tilde{{f}}\left( x \right)=\sum\limits_{j=1}^J {\sum\limits_{\Omega \in 
{\cal T}_{j} } {w_{j} \psi_{\Omega } \left( x \right)} } .
\end{equation}
The theory (see \cite{ED}) tells us that sparse approximation is achieved by ordering the wavelet components based on 
their norm
\begin{equation}
\label{eq7}
w_{j\left( {\Omega_{k_{1} } } \right)} \left\| {\psi_{\Omega_{k_{1} } } } 
\right\|_{p} \ge w_{j\left( {\Omega_{k_{2} } } \right)} \left\| {\psi 
_{\Omega_{k_{3} } } } \right\|_{p} \ge  \cdots 
\end{equation}
with the notation $\Omega \in {\cal T}_{j} \Rightarrow j\left( \Omega 
\right)=j$. Thus, the adaptive M-term approximation of a RF is
\begin{equation}
\label{eq8}
f_{M} \left( x \right):=\sum\limits_{m=1}^M {w_{j\left( {\Omega_{k_{m} } } 
\right)} \psi_{\Omega_{k_{m} } } \left( x \right)} .
\end{equation}
Observe that, contrary to most existing tree pruning techniques, where each tree 
is pruned separately, the above approximation process applies a `global' 
pruning strategy where the significant components can come from any node of 
any of the trees at any level. For simplicity, one could choose $w_{j} =1 
\mathord{\left/ {\vphantom {1 J}} \right. \kern-\nulldelimiterspace} J$, and 
obtain
\begin{equation}
\label{eq9}
f_{M} \left( x \right)=\frac{1}{J}\sum\limits_{m=1}^M {\psi_{\Omega_{k_{m} 
} } \left( x \right)} .
\end{equation}
\figurename~\ref{fig2} depicts an M-term (\ref{eq9}) selected 
from an RF ensemble. The red colored nodes illustrate the selection of the M 
wavelets with the highest norm values from the entire forest. Observe that 
they can be selected from any tree at any level, with no connectivity 
restrictions. 

\begin{figure}
\centering
\includegraphics[width=3in]{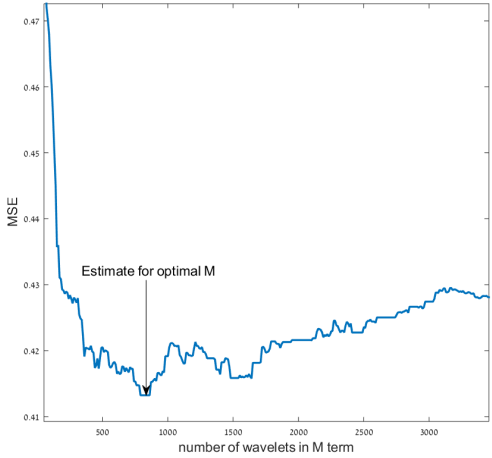}
\caption{``Red Wine Quality'' dataset - Numeric computation of M for optimal regression.}
\label{fig3}
\end{figure}

Figure \ref{fig3} depicts how the parameter $M$ is selected 
for the challenging ``Red Wine Quality'' dataset from the UCI repository 
\cite{UCI}. The generation of 10 decision trees on the 
training set creates approximately 3500 wavelets. The parameter M is then 
selected by minimization of the approximation error on an OOB validation set. In 
contrast with other pruning methods \cite{Lo}, using 
(\ref{eq7}), the wavelet approximation method may select significant components 
from any tree and any level in the forest. By this method, one does not need 
to predetermine the maximal depth of the trees and over-fitting is controlled 
by the selection of significant wavelet components. 

\section{Geometric multivariate function space theory}
\label{sec:mylabel2}
An important research area of approximation theory, pioneered by Pencho Petrushev, is the 
characterization of adaptive geometric approximation algorithms by 
generalizations of the classic `isotropic' Besov space to more `geometric' 
Besov-type spaces \cite{DDP}, \cite{DL}, \cite{KP}. We first review the definition and results of \cite{ED}. In essence, 
this is a generalization of a theoretical framework that has been successfully applied in the context of low dimensional
and structured signal processing \cite{De}, \cite{DJL}.
However, in the context of machine learning, we need to analyze unstructured and possibly high dimensional 
datasets.

Approximation Theory relates the sparsity of 
a function to its Besov smoothness index and supports cases where the function is not 
even continuous. 
For a function $f\in L_{\tau } \left( {\Omega } 
\right)$, $0<\tau \le \infty $, $h\in {\R}^{n}$ and $r\in {\N}$, we 
recall the $r$-th order difference operator
\begin{eqnarray*}
\Delta_{h}^{r} \left( {f,x} \right) &:=&\Delta_{h}^{r} \left( {f,\Omega ,x} \right) \\
                                    &:=& \sum\limits_{k=0}^r {\left( {-1} \right)^{r+k}\left( 
{{\begin{array}{*{20}c}
 r \hfill \\
 k \hfill \\
\end{array} }} \right)f\left( {x+kh} \right)},
\end{eqnarray*}
where we assume the segment $\left[ {x,x+rh} \right]$ is contained in $\Omega$. Otherwise,
we set the $\Delta_{h}^{r} \left( {f,\Omega ,x} \right)=0$. The \textbf{\textit{modulus of smoothness of order $r$}} is defined by
\[
\omega_{r} \left( {f,t} \right)_{\tau } :={\sup }_{\left| h 
\right|\le t} \left\| {\Delta_{h}^{r} \left( {f,\Omega ,\cdot } \right)} 
\right\|_{L_{\tau } \left( \Omega \right)} ,
\quad
t>0,
\]
where for $h\in {\R}^{n}$, $\left| h \right|$ denotes the norm of $h$. We 
also denote
\begin{equation} \label{mod-def}
\omega_{r} \left( {f,\Omega } \right)_{\tau } :=\omega_{r} \left( 
{f,\frac{diam\left( \Omega \right)}{r}} \right)_{\tau } .
\end{equation}
Next, we define the `weak-type' Besov smoothness of a function, subject to the geometry of a single (possibly adaptive) tree
\begin{defin}
\label{def-besov}
For $0<p<\infty $ and $\alpha >0$, we set $\tau =\tau \left( 
{\alpha ,p} \right)$, to be $1/\tau :=\alpha +1/p$. For a given function 
$f\in L_{p} \left( {\Omega_{0} } \right)$, $\Omega_{0} \subset {\R}^{n}$, 
and tree ${\cal T}$, we define the associated B-space smoothness in 
${\cal B}_{\tau }^{\alpha ,r} \left( {\cal T} \right)$, $r\in {\rm N}$, by
\begin{equation}
\label{eq10}
\left| f \right|_{{\cal B}_{\tau }^{\alpha ,r} \left( {\cal T} \right)} 
:=\left( {\sum\limits_{\Omega \in {\cal T}} {\left( {\left| \Omega 
\right|^{-\alpha }\omega_{r} \left( {f,\Omega } \right)_{\tau } } 
\right)^{\tau }} } \right)^{1/\tau },
\end{equation}
where, $\left| \Omega \right|$ denotes the volume of $\Omega$. 
\end{defin}

This notion of smoothness allows to handle functions that are not even continuous. The higher the index $\alpha$ for which (\ref{eq10})
is finite, the smoother the function is. This generalizes the Sobolev smoothness of differentiable functions that have their
partial derivatives integrable in some $L_{\tau}$ space. Also, the above definition generalizes the classical function space 
theory of Besov spaces, where the tree partitions are non-adaptive. In fact, classical Besov spaces are a special case, where the 
tree is constructed by partitioning into dyadic cubes, each time using $n$ levels of the tree. 
We recall that a `well clustered' function is in fact infinitely 
smooth in the right adaptively chosen Besov space.

\begin{lemma} \label{infi-besov-example}
 Let $f\left( x \right)=\sum\limits_{k=1}^K {P_{k} \left( x 
\right){\rm {\bf 1}}_{B_{k} } \left( x \right)} $, where each $B_{k} \subset 
\Omega_{0} $ is a box with sides parallel to the main axes and $P_{k} \in \Pi 
_{r-1} $. We further assume that $B_{k} \cap B_{j} =\emptyset $, whenever 
$j\ne k$. Then, there exists an adaptive tree partition ${\cal T}$, such 
that $f\in B_{\alpha }^{\alpha ,r} \left( {\cal T} \right)$, for any $\alpha 
>0$.
\end{lemma} 
\begin{IEEEproof} See \cite{ED}.
\end{IEEEproof} 

For a given forest 
${\cal F}=\left\{ {{\cal T}_{j} } \right\}_{j=1}^{J} $ and weights $w_{j} =1 
\mathord{\left/ {\vphantom {1 J}} \right. \kern-\nulldelimiterspace} J$, the 
$\alpha$- Besov semi-norm associated with the forest is
\begin{equation}
\label{eq11}
\left| f \right|_{{\cal B}_{\tau }^{\alpha ,r} \left( {\cal F} \right)} 
:=\frac{1}{J}\left( {\sum\limits_{j=1}^J {\left| f \right|_{{\cal B}_{\tau 
}^{\alpha ,r} \left( {{\cal T}_{j} } \right)}^{\tau } } } \right)^{1/\tau 
}.
\end{equation}

\begin{defin}
Given a (possibly adaptive) forest representation, we define the Besov smoothness index of $f$ by the maximal index 
$\alpha $ for which (\ref{eq11}) is finite. 
\end{defin}

\noindent \textbf{Remark} It is known 
that different geometric approximation schemes are characterized by 
different flavors of Besov-type smoothness. In this work, for example, all of our experimental results 
compute smoothness of representations using partitions along the main $n$ axes. This 
restriction may lead, in general, to potentially lower Besov smoothness of the 
underlying function and lower sparsity of the wavelet representation. Yet, the 
theoretical definitions and results of this paper can also apply to more 
generalized schemes where, for example, tree partitions are performed using arbitrary 
hyper-planes. In such a case, the smoothness index of a given function may 
increase.

Next, for a given tree ${\cal T}$ and parameter $0<\tau <p$, we denote the 
$\tau$-sparsity of the tree by 
\begin{equation}
\label{eq12}
N_{\tau } \left( {f,{\cal T}} \right)=\left( {\sum\limits_{\Omega \ne \Omega_0, \Omega \in {\cal 
T}} {\left\| {\psi_{\Omega } } \right\|_{p}^{\tau } } } \right)^{1/\tau }.
\end{equation}
Let us further denote the $\tau $-sparsity of a forest ${\cal F}$, by
\begin{align*}
\label{eq13}
 N_{\tau } \left( {f,{\cal F}} \right) & :=\frac{1}{J}\left( 
{\sum\limits_{j=1}^J {\sum\limits_{\Omega \ne \Omega_0, \Omega \in {\cal T}_{j} } {\left\| {\psi 
_{\Omega } } \right\|_{p}^{\tau } } } } \right)^{1/\tau } \\ 
& =\frac{1}{J}\left( {\sum\limits_{j=1}^J {N_{\tau } \left( {f,{\cal T}_{j} } 
\right)^{\tau }} } \right)^{1/\tau }.
\end{align*}
In the setting of a single tree constructed to represent a real-valued function and under mild 
conditions on the partitions (see remark after (\ref{eq3a}) and condition (\ref{sub-cond})) , the theory of 
\cite{DL} proves the equivalence
\begin{equation}
\label{eq14}
\left| f \right|_{{\cal B}_{\tau }^{\alpha ,r} \left( {\cal T} \right)} \sim 
N_{\tau } \left( {f,{\cal T}} \right).
\end{equation}
This implies that there are constants $0<C_{1} <C_{2} <\infty $, that depend 
on parameters such as $\alpha ,p,n,r$ and $\rho $ in condition (\ref{sub-cond}) below, such that
\[
C_{1} \left| f \right|_{{\cal B}_{\tau }^{\alpha ,r} \left( {\cal T} 
\right)} \le N_{\tau } \left( {f,{\cal T}} \right)\le C_{2} \left| f 
\right|_{{\cal B}_{\tau }^{\alpha ,r} \left( {\cal T} \right)} .
\]
Therefore, we also have for the forest model
\begin{equation}
\label{eq15}
\left| f \right|_{{\cal B}_{\tau }^{\alpha ,r} \left( {\cal F} \right)} \sim 
N_{\tau } \left( {f,{\cal F}} \right).
\end{equation}

In the setting in which we wish to apply our function theoretical approach, we are comparing smoothness of representation over 
different layers of DL networks. This implies that we are analyzing and comparing the smoothness a set of functions $f_k$, 
each over a different representation space of a different dimension $n_k$. This is, in some sense, non-standard in function space theory, where the space, or at least
the dimension, over which the functions have their domain is typically fixed. Specifically, observe that the equivalence (\ref{eq15})
depends on the dimension $n$ of the feature space. To this end, we add to the theory `dimension-free' analysis for the case
$r=1$. 
 
We begin with a Jackson-type estimate for the degree of the adaptive wavelet forest approximation, which we keep `dimension free'
for the case $r=1$.
\begin{thm}
\label{jackson-thm}
Let ${\cal F}=\left\{ {{\cal T}_{j} } \right\}_{j=1}^{J} $ be a 
forest. Assume there exists a constant $0<\rho <1$, such that for any domain 
$\Omega \in {\cal F}$ on a level $l$ and any domain ${\Omega }'\in {\cal 
F}$, on the level $l+1$, with $\Omega \cap {\Omega }'\ne \emptyset $, we 
have 
\begin{equation}
\label{sub-cond}
\left| {{\Omega }'} \right|\le \rho \left| \Omega \right|,
\end{equation}
where $\left| E \right|$ denotes the volume of $E\subset {\R}^{n}$.
For any $r \ge 1$, denote formally $f=\sum\limits_{\Omega \in {\cal F}} {w_{j\left( \Omega 
\right)} \psi_{\Omega } } $, and assume that $N_{\tau}(f,\cal F) <\infty $, where
\[
\frac{1}{\tau }=\alpha +\frac{1}{p}.
\]
Then, for the $M$-term approximation (\ref{eq8}) we have for $r=1$
\begin{equation}
\label{Jackson-r1}
\sigma_{M}(f) :=\left\| {f-f_{M} } \right\|_{p} \le C\left( {p,\alpha ,\rho} 
\right)J M^{-\alpha }N_{\tau}(f,\cal F).
\end{equation}
and for $r > 1$
\begin{equation}
\label{eq16}
\sigma_{M}(f) :=\left\| {f-f_{M} } \right\|_{p} \le C\left( {p,\alpha ,\rho,n } 
\right)J M^{-\alpha }N_{\tau}(f,\cal F).
\end{equation}
\end{thm}
\begin{IEEEproof} 
The proof in \cite{ED} shows (\ref{eq16}). To see (\ref{Jackson-r1}) we observe that the dimension $n$ comes into play in the
Nikolskii-type estimate for bounded convex domains $\Omega \subset \R^{n}$, and $r \ge 1$  
\[
\left\| \psi_{\Omega} \right\|_{\infty} \le c(p,n,r) |\Omega|^{-1/p} \left\| \psi_{\Omega} \right\|_{p}.
\]
However, for the special case of $r=1$ this actually simplifies to
\[
\left\| \psi_{\Omega} \right\|_{\infty} = |\Omega|^{-1/p} \left\| \psi_{\Omega} \right\|_{p}.
\]
\end{IEEEproof} 
Using the equivalence (\ref{eq15}), we get for any $r \ge 1$
\[
\sigma_{M}(f) \le C\left( {p,\alpha ,\rho,n } \right)J M^{-\alpha }\left| f \right|_{{\cal B}_{\tau }^{\alpha ,r} \left( {\cal F} \right)},
\]
which is not a `dimension-free' Jackson estimate, as the one will show below for $r=1$ (see (\ref{Jack-free})). Next, we present a simple invariance property 
of the smoothness analysis under higher dimension embedding. 

\begin{lemma} \label{besov-inv}
Let $\{x_i\}$, $x_i \in [0,1]^n$, with values $f(x_i)\in \R^{L-1}$, $i \in I$. Let $\cal F$ be a forest
approximation of the data. For any $m \ge 0$, let $\{ \tilde{x}_i \}$ be defined by $\tilde{x}_i = (x_i,0,...,0) \in [0,1]^{n+m}$, $i \in I$. 
Let us further define $\tilde{f}(\tilde{x}_i):= f(x_i)$. Next, denote by $\tilde{\cal F}$ a forest defined over $[0,1]^{n+m}$ which is the natural extension of $\cal F$, using the same trees with same partitions over the first $n$ dimensions.
Then, for $r=1$ and any $\tau >0$, $N_{\tau }, \left( {\tilde{f},{\tilde{\cal F}}} \right)=N_{\tau } \left( {f,{\cal F}} \right)$. 
\end{lemma}
\begin{IEEEproof} 
Let $\Omega' \in {\cal F}$ be the domains of the trees of $\cal F$, with wavelets of the type 
\[
\psi_{{\Omega }'}(x) ={\rm {\bf 1}}_{{\Omega }'}(x) \left( {\vec{{E}}_{{\Omega 
}'} -\vec{{E}}_{\Omega } } \right).
\]
Recall that $N_{\tau } \left( {f,{\cal F}} \right)$ is the $l_{\tau}$ norm of the sequence of the wavelet
norms given by (\ref{eq5}). 

Now, for each domain $\Omega' \in {\cal F}$ and the corresponding domain $\tilde{\Omega'} \in \tilde{\cal F}$, the normalization
of the feature space into $[0,1]^n$ and the higher dimensional embedding in $[0,1]^{n+m}$ ensures that
\[
\left| \Omega' \right| = \left| \Omega' \right| \times  \left| {[0,1]^m} \right| = \left| \tilde{\Omega'} \right|.
\]
Since the vector means $\{ {\vec{{E}}_{{\Omega }'}} \}$ remain unchanged under the higher dimensional embedding, we have 
\[
\left\| {\psi_{{\Omega }'} } \right\|_{L_p[0,1]^n} = \left\| {\psi_{\tilde{\Omega '}} } \right\|_{L_p[0,1]^{n+m}}.
\]
This gives $N_{\tau } \left( {\tilde{f},{\tilde{\cal F}}} \right)=N_{\tau } \left( {f,{\cal F}} \right)$. 
\end{IEEEproof} 

Next, to allow our smoothness analysis to be `dimension free' we modify the modulus of smoothness (\ref{mod-def}) for $r=1$ and use the following 
form of `averaged modulus' 
\begin{defin} For a function $f:\Omega \rightarrow \R^{L-1}$ we define
\begin{equation} \label{aver-mod}
{{w}_{1}}{{\left( f,\Omega  \right)}_{\tau }}:={{\left( \int_{\Omega }{\left\| f\left( x \right)-{{{\vec{E}}}_{\Omega }} \right\|_{{{l}_{2}}\left( L-1 \right)}^{\tau }dx} \right)}^{1/\tau }},
\end{equation}
where ${\vec{E}}_{\Omega }$ is the average of $f$ over $\Omega$.
\end{defin}
It is well known that averaged forms of the modulus are equivalent to the form (\ref{mod-def}), but with constants that depend on the dimension. 
However, replacing (\ref{mod-def}) with (\ref{aver-mod}) allows us to produce `dimension-free' analysis. We use (\ref{aver-mod}) to define
\[
{{\left| f \right|}_{\tilde{B}_{\tau }^{\alpha ,1}\left( \mathcal{T} \right)}}:={{\left( \sum\limits_{\Omega \in \mathcal{T}}{{{\left( {{\left| \Omega  \right|}^{-\alpha }}{{w}_{1}}{{\left( f,\Omega  \right)}_{\tau }} \right)}^{\tau }}} \right)}^{1/\tau }}.
\]
We can now show
\begin{thm} \label{dim-free-equiv} Let $f:\Omega_0 \rightarrow \R^{L-1}$. Then the following equivalence holds for the case $r=1$, 
\[
{{\left| f \right|}_{\tilde{B}_{\tau }^{\alpha ,1}\left( \mathcal{F} \right)}}\sim {{N}_{\tau }}\left( f,\mathcal{F} \right),
\]
where ${1}/{\tau }\;=\alpha +1/p$, and the constants of equivalence depend on $\alpha, \tau, \rho$, but not $n$. 
\end{thm} 
\begin{IEEEproof} 
See the Appendix
\end{IEEEproof} 
This equivalence together with (\ref{Jackson-r1}) imply that for $r=1$ we do have a `dimension-free' Jackson estimate
\begin{equation} \label{Jack-free}
\sigma_{M}(f) \le C\left( {p,\alpha ,\rho} \right)J M^{-\alpha }\left| f \right|_{{\tilde{\cal B}}_{\tau }^{\alpha ,1} \left( {\cal F} \right)}.
\end{equation}

\section{Smoothness analysis of the representation layers in deep learning networks}
\label{sec:smoothAnalysis}

We now explain how the theory presented in Section \ref{sec:mylabel2} is used to estimate the `weak-type' smoothness
of a given function in a given representation layer. Recall from the introduction that we create a representation of images at  
layer $0$ by concatenating the $\sqrt{n_0}$
rows of pixel values of each grayscale image, to create a vector of dimension $n_0$ (or $3 \times n_0$ for a color image). We also normalize the pixel
values to the range $[0,1]$. We then transform the class labels into vector-values in the space $\R^{L-1}$ by assigning 
each label to a vertex of a standard simplex (see Section \ref{sec:overview}). Thus, the images are considered as samples of a function
$f_0:[0,1]^{n_0} \rightarrow \R^{L-1}$. In the same manner, we associate with each $k$-th layer of a DL network, a function 
$f_k:[0,1]^{n_k} \rightarrow \R^{L-1}$, where $n_k$ is the number of features/neurons at the $k$-th layer. The samples of $f_k$ are obtained by applying the network on the original images 
up the given $k$-th layer. For example, in a convolution layer, we capture the representations after the cycle of convolution, non-linearity and pooling. We then extract vectors created by normalizing and concatenating the feature map values corresponding to the images. Recall that although the functions $\{f_k\}$ are embedded in different dimensions $\{n_k\}$,
through the simple normalizing of the features, our method is able to assign smoothness indices to each layer that are comparable. 
 
Next we describe how we estimate the smoothness of each function $f_k$. To this end, we have made several improvements and simplifications to the method of \cite{ED}. We compute a RF over the samples of $f_k$ with the choice $r=1$ and
then apply the wavelet decomposition of the RF (see Section \ref{sec:overview}). For each $k$ and $M$ one computes the discrete
error of the wavelet $M$-term approximation for the case $p=2$
\begin{equation} \label{detail-S-M}
\sigma_{M}(f_k)^2=\frac{1}{|I|}\sum_{i\in I} \|S_M(f_k)(x_i)-f_k(x_i) \|_{l_2(L-1)}^2.
\end{equation}
We then use the theoretical estimate (\ref{Jack-free}) and numeric estimate of $\sigma_{M}:=\sigma_{M}(f_k)$ in (\ref{detail-S-M}) to model the error function by $\sigma_{M} \sim c_k M^{-\alpha_k}$
for unknown $c_k,\alpha_k$. Notice that the constant $c_k$ absorbs the terms relating to the absolute constant, the number of trees in RF model as well as the Besov-norm. Numerically, one simply models $log(\sigma_{M}) \sim log(c_k) - \alpha log(M)$, $M=1,...,\tilde{M}$,
and then solves through least squares for $c_k, \alpha_k$. Finally, we set $\alpha_k$ as our estimate for the `critical' Besov smoothness index of $f_k$. 

\noindent {\bf Remarks}:

\noindent (i) Observe that for the fit of $c_k$ and $\alpha_k$, we only use $\tilde{M}$ significant terms, so as to avoid fitting the `noisy' tail of the exponential expression. In some cases, we allow ourselves to select $\tilde{M}$ adaptively, by discarding a tail of wavelet
components that is over-fitting the training data, but increasing the error on validation set samples (see Figure \ref{fig3}). However, in cases where the goal is to demonstrate understanding of generalization
we restrict the analysis to only using the training set and then pre-select $\tilde{M}$ (e.g. $\tilde{M}=1000$ in the experiments we review below).

\noindent (ii) Notice that since 
each representation space can be of very different dimension, it is crucial that the method is invariant under different
dimension embedding. 

\noindent (iii) We note that
this approach to compute the geometric Besov smoothness of a labeled dataset is a significant generalization of the method used in \cite{DJL} to compute the (classical) Besov smoothness of a single image. Nevertheless, there is a distinct similar 
underlying function space approach.

\section{Applications and Experimental Results }
\label{sec:applications}
In all of the experiments we used TensorFlow networks models. we extracted the representation of any data sample (e.g. image) in any layer of a network, by simply running the TensorFlow `Session' object with
the given layer and the data sample as the parameters. 

The computation of the Besov smoothness index
in a given feature space is implemented as explained in Section \ref{sec:smoothAnalysis}. We used an updated version of the code of \cite{ED} which is available via the link 
in \cite{WRF}. For the hyper-parameter that determines the number of $M$-term errors (\ref{detail-S-M}), $M=1,...,\tilde{M}$, which are used to model the $\alpha$-smoothness, we used $\tilde{M} = 1000$. The code was executed on the Amazon Web Services cloud, on r3.8xlarge configurations that have 32 virtual CPUs and 244 GB of memory. We note that computing the smoothness of all the representation of a certain dataset of images over all layers requires
significant computation. One needs to create a RF approximation of the representation at each layer, sort the wavelet components based on their norms and compute the errors $M$-term errors (\ref{detail-S-M}), $M=1,...,\tilde{M}$,
before the numeric fit of the smoothness index can be computed. Thus, in our experiments, we computed and used for the smoothness fit only the errors $\sigma_{10i}$, $i=1,...,100$, to speed up the computation.

\subsection{Smoothness analysis across deep learning layers}
\label{subsec:layers}

\begin{figure}
\centering
\includegraphics[width=3.5in]{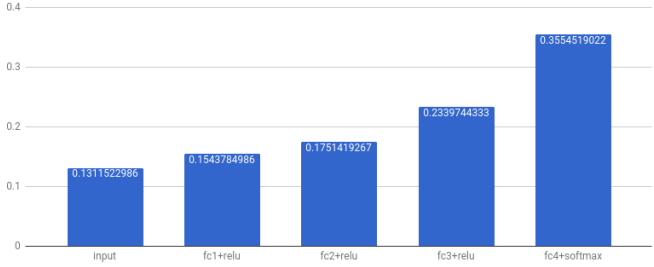}
\caption{Smoothness analysis of the layer representations of ``Urban8K'' using the DeepListen \cite{DeepListen} fully-connected architecture}
\label{fig:Urban-layers}
\end{figure}

We now present results for estimates of smoothness analysis in layer representations for some datasets and trained networks. We begin with the audio dataset ``Urban Sound Classification'' from \cite{Urban}.
We applied our smoothness analysis on representations of the ``Urban8K'' audio data at the layers of the DeepListen model \cite{DeepListen}
which achieves an accuracy of $61\%$. The network is a simple feed-forward of 4 fully connected layers with ReLU non-linearities. As described in 
Section \ref{sec:smoothAnalysis}, we created a functional representation of the data at each layer and estimated 
the Besov $\alpha$ smoothness index at each layer. In Figure \ref{fig:Urban-layers} we see how the clustering is `unfolded' by the network, as the Besov $\alpha$-index increases from layer to layer.

\begin{figure}
\centering
\includegraphics[width=3.5in]{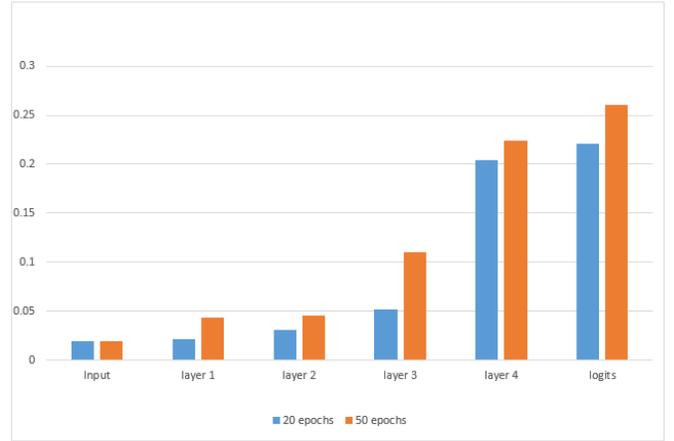}
\caption{Smoothness analysis of DL layers representations of CIFAR10}
\label{fig:CIFAR}
\end{figure}

Next we present results on image datasets. We trained the network \cite{TensorAlexa} on the CIFAR10 image dataset \cite{Cif}. As described in \cite{TensorAlexa}, the images
were cropped to size $24\times 24$. The network has 2 convolution layers (with 9216 and 2304 features, respectively) and 2 fully connected layers (with 384 and 192 features, respectively) with an additional 
soft-max layer (with final layer `logits' of 10 classes). The training set data size is 50,000 and the testing 10,000. As expected \cite{TensorAlexa}, the 
trained network achieves $ 86\%$ accuracy on the testing data. In Figure \ref{fig:CIFAR} we see a clear indication of how the 
smoothness begins to evolve during the training after 20 epochs and the `unfolding' of the clustering improves from layer to layer. We also see that the 
smoothness improves after 50 epochs, correlating with the improvement of the accuracy.

\begin{figure}
\centering
\includegraphics[width=3.5in]{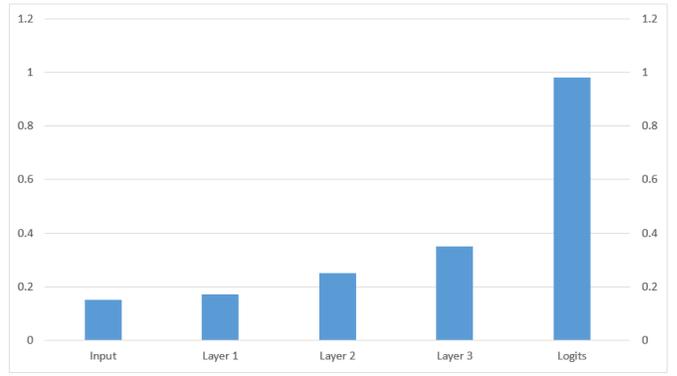}
\caption{Smoothness analysis of DL layers representations of MNIST}
\label{fig:MNIST}
\end{figure}

We now describe our experiments with the well-known MNIST dataset of 60,000 training and 10,000 testing images  \cite{MNIST-dat}. The DL network configuration we used is the `textbook' version of \cite{MNIST-Ten}, which is composed of two convolution layers and two fully connected layers.
Training the model for 100 epochs produce a model with $99.51\%$ accuracy on the training data and a clear monotone increase of Besov smoothness  across layers as shown in Figure \ref{fig:MNIST}.

\subsection{Smoothness analysis of mis-labeled datasets}
\label{subsec:mislabeling}

Following \cite{Zhang}, we applied random mis-labeling to the MNIST and CIFAR10 image sets at various levels. We randomly picked subsets
of size $q\%$ of the size of dataset, with $q=10\%, 20\%, 30\%, 40\%$, and then for each image in this subset we picked a random label. We then trained the network of \cite{MNIST-Ten} on the misclassified MNIST datasets and the network of \cite{TensorAlexa} on the misclassified CIFAR10 set. We emphasize that
the goal of this experiment is to understand generalization \cite{Zhang} and automatically detect the level of corruption solely from the 
smoothness analysis of the training data. Recall from \cite{Zhang} that a network can converge relatively quickly to an over-fit even on 
highly mis-classified training sets. Thus, convergence is not a good indication to the generalization capabilities and specifically to the 
level of mis-labeling in the training data. 

\begin{figure}
\centering
\includegraphics[width=3.5in]{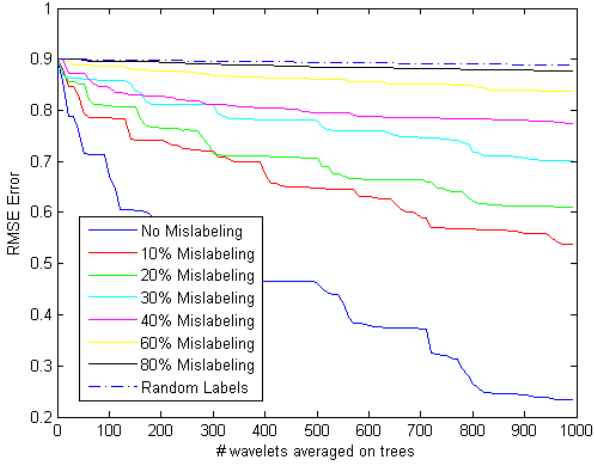}
\caption{Precision error decay with adaptive wavelet approximation on mis-labeled MNIST}
\label{fig:noise-and-error}
\end{figure}

Next, we created a wavelet decomposition of RF on the representation of the training set at the last inner layer of the network. Typically, this is the fully connected layer right before the softmax.
In Figure \ref{fig:noise-and-error} we see the decay of the precision error of the adaptive wavelet approximations (\ref{eq9}) as we add 
more wavelet terms. It is clear that datasets with less mis-labeling have more `sparsity', i.e., are better approximated with less wavelet terms. We also measured for each mis-labeled training dataset the Besov $\alpha$-smoothness at the last inner layer. The results are presented in Table \ref{tab:alpha-noise}. We see a strong correlation between the amount of mis-labeling and the smoothness. 

\begin{table}
\begin{centering}
\begin{tabular}{|c|c|c|c|c|c|c|c|}
\hline 
\bf{Mis-labeling} & 0\%  & 10\%  & 20\%  & 30\% & 40\%   \tabularnewline
\hline  
\bf{MNIST smoothness} & 0.28 & 0.106 & 0.084 & 0.052 & 0.03  \tabularnewline
\hline 
\bf{CIFAR10 smoothness} & 0.204 & 0.072 & 0.053 & 0.051 & 0.003  \tabularnewline
\hline 
\end{tabular}
\par\end{centering}
\caption{Smoothness analysis of mis-labeled image images}
\label{tab:alpha-noise}
\end{table}

\section{Conclusion} \label{sec:conc}

In this paper we presented a theoretical approach to the analysis of the performance of the hidden layers in DL architectures. 
We plan to continue the experimental analysis of deeper architectures and larger datasets (see e.g. \cite{SSSG}) and hope
to demonstrate that our approach is applicable to a wide variety of machine learning and deep learning
architectures. As some advanced DL architectures have millions of features in their 
hidden layers, we will need to overcome the problem of estimating representation smoothness in such high dimensions. Furthermore,
in some of our experiments we noticed interesting phenomena within sub-components of the layers (e,g, the different operations of convolution, non-linearity and pooling).
We hope to reach some understanding and share some insights regarding these aspects too.

\appendix[Proof of Theorem \ref{dim-free-equiv}] 

%
Obviously, it is sufficient to prove the equivalence for a single tree $\mathcal{T}$. Observe that condition (\ref{sub-cond})
also implies that for any $\Omega' \in \mathcal{T}$, with parent $\Omega$, we also have
\[
\left| \Omega \right| \le (1-\rho)^{-1}\left|{\Omega'}\right|.
\]
We use this as well as (\ref{eq4}) to prove the first direction of the equivalence as follows
\begin{align*}
  & {{N}_{\tau }}{{\left( f,\mathcal{T} \right)}^{\tau }}=\sum\limits_{{\Omega }'\ne {{\Omega }_{0}},{\Omega }'\in \mathcal{T}}{\left\| {{\psi }_{{{\Omega }'}}} \right\|_{p}^{\tau }} \\ 
 & =\sum\limits_{\begin{smallmatrix} 
 {\Omega }'\ne {{\Omega }_{0}},{\Omega }'\in \mathcal{T}, \\ 
 \text{ }\Omega \text{ parent of }{\Omega }' 
\end{smallmatrix}}{{{\left( {{\left| {{\Omega }'} \right|}^{1/p}}{{\left\| {{{\vec{E}}}_{{{\Omega }'}}}-{{{\vec{E}}}_{\Omega }} \right\|}_{{{l}_{2}}\left( L-1 \right)}} \right)}^{\tau }}} \\ 
 & =\sum\limits_{{\Omega }'\ne {{\Omega }_{0}},{\Omega }'\in \mathcal{T}}{{{\left( {{\left| {{\Omega }'} \right|}^{1/p-1/\tau }}{{\left\| {{\psi }_{{{\Omega }'}}} \right\|}_{\tau }} \right)}^{\tau }}} \\ 
 & \le c\left( \tau  \right)\sum\limits_{\begin{smallmatrix} 
 {\Omega }'\ne {{\Omega }_{0}},{\Omega }'\in \mathcal{T} \\ 
 ,\Omega \text{ parent of }{\Omega }' 
\end{smallmatrix}}{\left\{ {{\left( {{\left| {{\Omega }'} \right|}^{-\alpha }}{{\left\| {{\left\| f\left( \cdot  \right)-{{{\vec{E}}}_{{{\Omega }'}}} \right\|}_{{{l}_{2}}\left( L-1 \right)}} \right\|}_{{{L}_{\tau }}\left( {{\Omega }'} \right)}} \right)}^{\tau }} \right.} \\ 
 & \qquad +\left. {{\left( {{\left| {{\Omega }'} \right|}^{-\alpha }}{{\left\| {{\left\| f\left( \cdot  \right)-{{{\vec{E}}}_{\Omega }} \right\|}_{{{l}_{2}}\left( L-1 \right)}} \right\|}_{{{L}_{\tau }}\left( {{\Omega }'} \right)}} \right)}^{\tau }} \right\} \\ 
 & \le c\left( \tau ,\rho, \alpha  \right)\sum\limits_{\begin{smallmatrix} 
 \Omega \ne {{\Omega }_{0}},,{\Omega }'\in \mathcal{T} \\ 
 ,\Omega \text{ parent of }{\Omega }' 
\end{smallmatrix}}{\left\{ {{\left( {{\left| {{\Omega }'} \right|}^{-\alpha }}{{\left\| {{\left\| f\left( \cdot  \right)-{{{\vec{E}}}_{{{\Omega }'}}} \right\|}_{{{l}_{2}}\left( L-1 \right)}} \right\|}_{{{L}_{\tau }}\left( {{\Omega }'} \right)}} \right)}^{\tau }} \right.} \\ 
 & \left. \qquad  +{{\left( {{\left| \Omega  \right|}^{-\alpha }}{{\left\| {{\left\| f\left( \cdot  \right)-{{{\vec{E}}}_{\Omega }} \right\|}_{{{l}_{2}}\left( L-1 \right)}} \right\|}_{{{L}_{\tau }}\left( \Omega  \right)}} \right)}^{\tau }} \right\} \\ 
 & \le 2c\left( \tau ,\rho, \alpha  \right)\sum\limits_{\Omega \in \mathcal{T}}{{{\left( {{\left| \Omega  \right|}^{-\alpha }}{{\left\| {{\left\| f\left( \cdot  \right)-{{{\vec{E}}}_{\Omega }} \right\|}_{{{l}_{2}}\left( L-1 \right)}} \right\|}_{{{L}_{\tau }}\left( \Omega  \right)}} \right)}^{\tau }}} \\ 
 & =2c\left( \tau ,\rho, \alpha  \right)\sum\limits_{\Omega \in \mathcal{T}}{{{\left( {{\left| \Omega  \right|}^{-\alpha}}{{w}_{1}}{{\left( f,\Omega  \right)}_{\tau }} \right)}^{\tau }}} \\ 
 & =c\left| f \right|_{\tilde{B}_{\tau }^{\alpha ,1}}^{\tau }.  
\end{align*}
We now prove the other direction. We assume $0 < \tau \le 1$ (the case $1 < \tau < \infty$ is similar). 
For any $\Omega \in \mathcal{T}$ we have

\begin{equation} \label{aux1}
 {{w}_{1}}\left( f,\Omega  \right)_{\tau }^{\tau }\le \sum\limits_{{\Omega }'\in \mathcal{T},{\Omega }'\subset \Omega }{\left\| {{\psi }_{{{\Omega }'}}} \right\|_{\tau }^{\tau }}, 
\end{equation}
\noindent by the following estimates
\begin{align*}
  & {{w}_{1}}\left( f,\Omega  \right)_{\tau }^{\tau }=\int_{\Omega }{\left\| \sum\limits_{{\Omega }'\in \mathcal{T}}{{{\psi }_{{{\Omega }'}}}\left( x \right)}-{{{\vec{E}}}_{\Omega }} \right\|_{{{l}_{2}}\left( L-1 \right)}^{\tau }dx} \\ 
 & \qquad =\int_{\Omega }{\left\| \sum\limits_{{\Omega }'\in \mathcal{T}}{{{\psi }_{{{\Omega }'}}}\left( x \right)}-\sum\limits_{{\Omega }'\in \mathcal{T},\Omega \subseteq {\Omega }'}{{{\psi }_{{{\Omega }'}}}\left( x \right)} \right\|_{{{l}_{2}}\left( L-1 \right)}^{\tau }dx} \\ 
 & \qquad =\int_{\Omega }{\left\| \sum\limits_{{\Omega }'\in \mathcal{T},{\Omega }'\subset \Omega }{{{\psi }_{{{\Omega }'}}}\left( x \right)} \right\|_{{{l}_{2}}\left( L-1 \right)}^{\tau }dx} \\ 
 & \qquad \le \sum\limits_{{\Omega }'\in \mathcal{T},{\Omega }'\subset \Omega }{\left\| {{\psi }_{{{\Omega }'}}} \right\|_{\tau }^{\tau }}. 
\end{align*}
Also, observe that by condition (\ref{sub-cond}), for any $\Omega' \in \mathcal{T}$
\begin{equation} \label{aux2}
\sum\limits_{\Omega \in \mathcal{T},{\Omega }'\subset \Omega }{{{\left( \frac{\left| {{\Omega }'} \right|}{\left| \Omega  \right|} \right)}^{\alpha \tau }}}\le \sum\limits_{k=1}^{\infty }{{{\rho }^{k\alpha \tau }}}\le c\left( \rho ,\alpha ,\tau  \right).
\end{equation}
We apply (\ref{aux1}) and (\ref{aux2}) to conclude 
\begin{align*}
  & \left| f \right|_{\tilde{B}_{\tau }^{\alpha ,1}\left( \mathcal{T} \right)}^{\tau }\le \sum\limits_{\Omega \in \mathcal{T}}{{{\left| \Omega  \right|}^{-\alpha \tau }}\sum\limits_{{\Omega }'\in \mathcal{T},{\Omega }'\subset \Omega }{\left\| {{\psi }_{{{\Omega }'}}} \right\|_{\tau }^{\tau }}} \\ 
 & =\sum\limits_{\Omega \in \mathcal{T}}{\sum\limits_{{\Omega }'\in \mathcal{T},{\Omega }'\subset \Omega }{{{\left( \frac{\left| {{\Omega }'} \right|}{\left| \Omega  \right|} \right)}^{\alpha \tau }}{{\left( {{\left| {{\Omega }'} \right|}^{-\alpha }}{{\left\| {{\psi }_{{{\Omega }'}}} \right\|}_{\tau }} \right)}^{\tau }}}} \\ 
 & =\sum\limits_{\Omega' \ne \Omega_0, {\Omega }'\in \mathcal{T}}{{{\left( {{\left| {{\Omega }'} \right|}^{-\alpha }}{{\left\| {{\psi }_{{{\Omega }'}}} \right\|}_{\tau }} \right)}^{\tau }}\sum\limits_{\Omega \in \mathcal{T},{\Omega }'\subset \Omega }{{{\left( \frac{\left| {{\Omega }'} \right|}{\left| \Omega  \right|} \right)}^{\alpha \tau }}}} \\ 
 & \le c\left( \alpha ,\tau ,\rho  \right)\sum\limits_{\Omega' \ne \Omega_0, {\Omega }'\in \mathcal{T}}{{{\left( {{\left| {{\Omega }'} \right|}^{-\alpha }}{{\left| {{\Omega }'} \right|}^{1/\tau -1/p}}{{\left\| {{\psi }_{{{\Omega }'}}} \right\|}_{p}} \right)}^{\tau }}} \\ 
 & =c\left( \alpha ,\tau ,\rho  \right)\sum\limits_{\Omega' \ne \Omega_0, {\Omega }'\in \mathcal{T}}{\left\| {{\psi }_{{{\Omega }'}}} \right\|_{p}^{\tau }} \\
 & =c{{N}_{\tau }}{{\left( f,\mathcal{T} \right)}^{\tau }}.  
\end{align*}

\ifCLASSOPTIONcompsoc
  \section*{Acknowledgments}
\else
  \section*{Acknowledgment}
\fi

The authors would like to thank Vadym Boikov, WIX AI and Kobi Gurkan, Tel-Aviv University, for their help with running the experiments. This research was carried out with the 
generous support of the Amazon AWS Research Program. 

\ifCLASSOPTIONcaptionsoff
  \newpage
\fi

\end{document}